\documentclass[sigconf]{acmart}

\usepackage{graphicx}%
\usepackage{multirow}%

\AtBeginDocument{%
  }

\setcopyright{acmlicensed}
\copyrightyear{2024}
\acmYear{2024}

\acmConference[1st Workshop on GenAI and RAG Systems for Enterprise]{CIKM 2024}{October 24, 2024}{Boise, Idaho, USA}

\begin{document}

\title{RAG based Question-Answering for Contextual Response Prediction System}

\author{Sriram Veturi}
\authornote{Both first authors contributed equally to this research.}
\email{sriram_veturi@homedepot.com}
\author{Saurabh Vaichal}
\authornotemark[1]
\email{saurabh_s_vaichal@homedepot.com}
\affiliation{%
  \institution{The Home Depot}
  \city{Atlanta}
  \state{Georgia}
  \country{USA}
}
\author{Reshma Lal Jagadheesh}
\authornote{Work completed during employment at the Home Depot}
\affiliation{%
  \institution{The Home Depot}
  \city{Atlanta}
  \state{Georgia}
  \country{USA}
}

\author{Nafis Irtiza Tripto}
\authornote{Work completed during internship at The Home Depot}
\affiliation{%
  \institution{The Pennsylvania State University}
  \country{USA}}
\email{nit5154@psu.edu}

\email{reshma_lal_jagadheesh@homedepot.com}

\author{Nian Yan}
\affiliation{%
 \institution{The Home Depot}
 \city{Atlanta}
  \state{Georgia}
 \country{USA}
 }
 \email{nian_yan@homedepot.com}

\renewcommand{\shortauthors}{Veturi \& Vaichal et al, 2024}


\begin{abstract}
Large Language Models (LLMs) have shown versatility in various Natural Language Processing (NLP) tasks, including their potential as effective question-answering systems. However, to provide precise and relevant information in response to specific customer queries in industry settings, LLMs require access to a comprehensive knowledge base  to avoid hallucinations. Retrieval Augmented Generation (RAG) emerges as a promising technique to address this challenge. Yet, developing an accurate question-answering framework for real-world applications using RAG entails several challenges: 1) data availability issues, 2) evaluating the quality of generated content, and 3) the costly nature of human evaluation. In this paper, we introduce an end-to-end framework that employs LLMs with RAG capabilities for industry use cases. Given a customer query, the proposed system retrieves relevant knowledge documents and leverages them, along with previous chat history, to generate response suggestions for customer service agents in the contact centers of a major retail company. Through comprehensive automated and human evaluations, we show that this solution outperforms the current BERT-based algorithms in accuracy and relevance. Our findings suggest that RAG-based LLMs can be an excellent support to human customer service representatives by lightening their workload.
\end{abstract}

%




\maketitle

\section{Introduction}

\begin{figure}
    \includegraphics[width=0.5\textwidth]{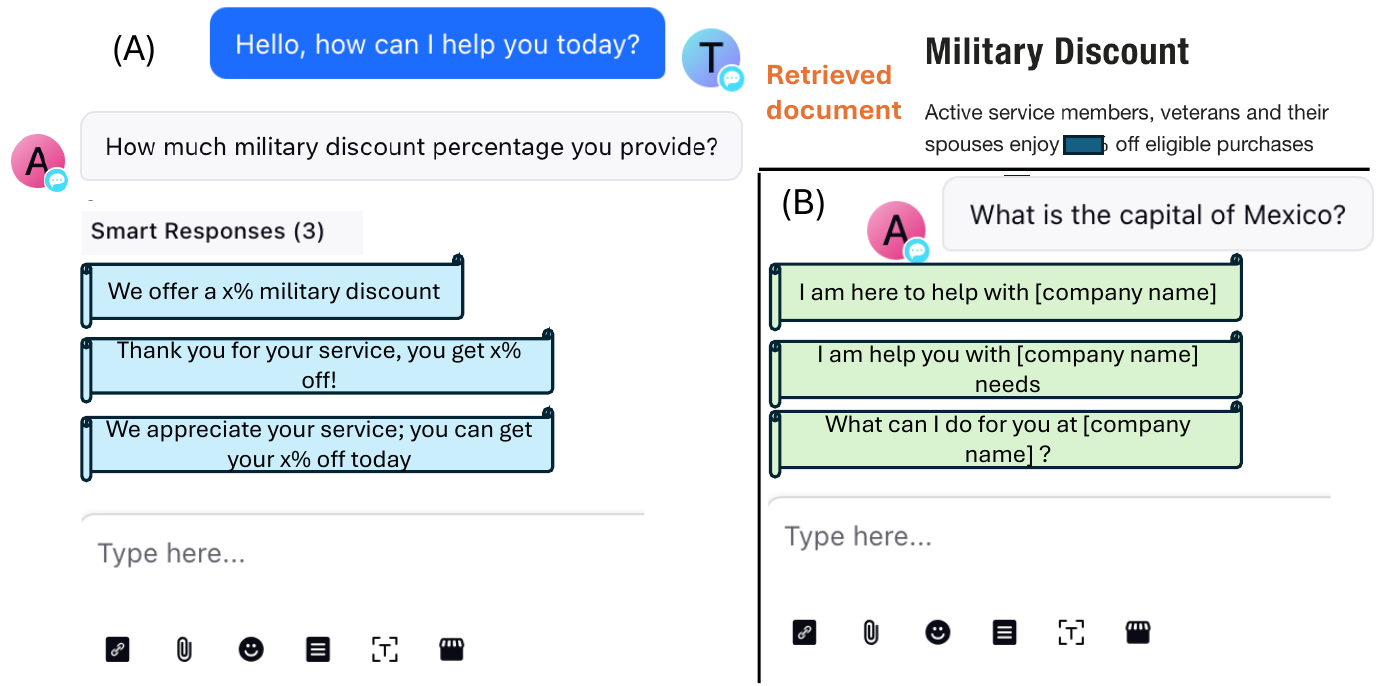}
    \caption{Example of the Response Prediction System. \textbf{(A)}: For a valid query, the system retrieves the relevant document and proposes the appropriate responses from where the agent choose. \textbf{(B)}: For an out-of-domain query, it guides the user to ask a relevant question.}
     \label{fig:SmartReply-diag}
\end{figure}

With the advent of ChatGPT and similar tools in mainstream media, Large Language Models (LLMs) have emerged as the standard solution for addressing a wide range of language understanding tasks. However, they can generate incorrect or biased information \cite{tonmoy2024comprehensive}, as their responses are based on patterns learned from data that may not always contain necessary knowledge in a close domain. To address this issue, Retrieval Augmented Generation (RAG) \cite{lewis2021retrievalaugmented} is commonly used to ground LLMs in factual information. The RAG architecture processes user input by first retrieving a set of documents similar to the query, which the language model then uses to generate a final prediction. While RAG-based architectures have been successful in various open-domain question answering (Q/A) tasks \cite{siriwardhana2023improving, zhao2024retrieval, izacard2020leveraging}, limited research has explored their scaling dynamics in real conversational scenarios. Therefore, our research is one of the pioneering efforts in exploring the feasibility of an RAG-based approach for developing a knowledge-grounded response prediction system specifically tailored for the contact center of a major retail company.

\begin{figure*}[h]
    \includegraphics[width=\textwidth]{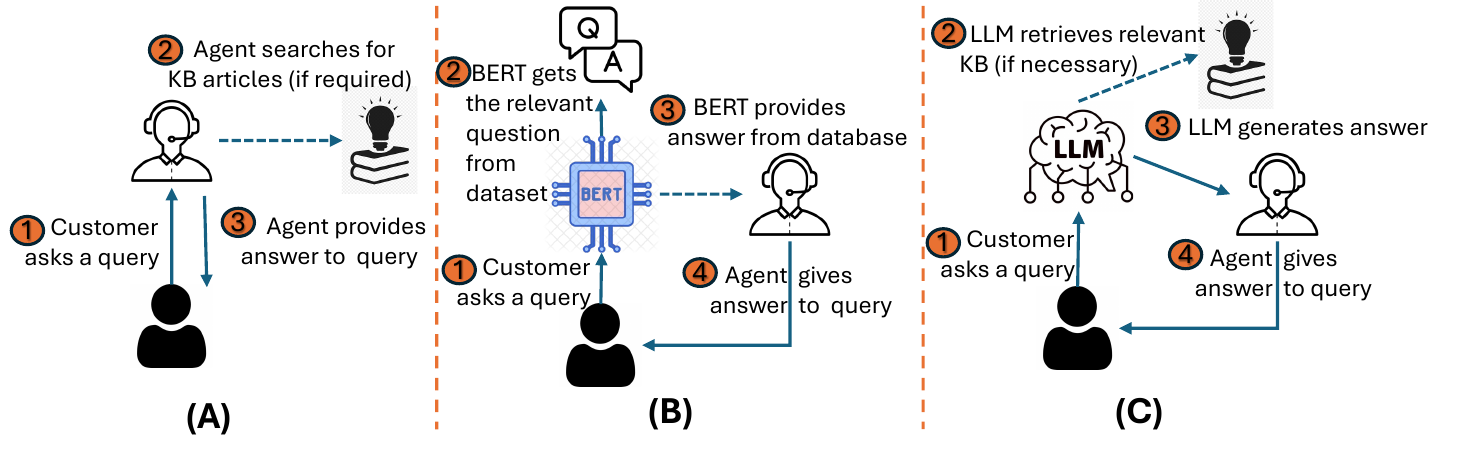}
    \caption{Overview of the systems: (A) Agents respond to queries by manually searching for relevant documents, (B) The existing BERT-based system, which extracts relevant Q/A pairs from the given query and provides suggested answers to the agents, (C) The proposed RAG LLM system, where the LLM retrieves relevant KB articles (if necessary) and generates answers based on the query and the retrieved articles.}
    \label{fig:system_overview}
\end{figure*}

LLMs have recently been widely adopted across various industries, particularly in contact centers, to enhance chatbot development and agent-facing automation \cite{wortmann2020modeling,chiang2024chatbot,freire2024chatbots}. A prime example is the Response Prediction System (RPS), an agent-assist solution that generates contextually relevant responses, enabling agents to efficiently address customer queries with a single click. This boosts productivity, improves customer experience, and streamlines communication processes. In industry settings, the focus is on generating accurate, contextually appropriate responses with minimal latency. Therefore, RAG-based responses, grounded in company policies, deliver swift and accurate resolutions to customer issues. Figure \ref{fig:SmartReply-diag} demonstrates a possible example of RPS in real settings, where the agent can directly utilize the generated response with a single click.

However, implementing RAG for industry-specific use cases to assist human agents in generating valid responses involves several architectural decisions that can affect performance and viability. The retrieval style can be integrated into both encoder-decoder \cite{yu-2022-retrieval,izacard2022atlas}) and decoder-only models \cite{khandelwal2020generalization,borgeaud2022improving,shi2022knnprompt,rubin2022learning}, with various embedding and prompting techniques influencing the final LLM output. In contact centers, where the risk of hallucinations is high and can critically impact business performance, ReAct (Reason+Act) \cite{yao2023react} prompts can help mitigate issues. Therefore, our research focuses on developing an optimal RAG based knowledge-grounded RPS for a major retail company's contact center.  To ensure response accuracy, we also conduct thorough evaluations with human evaluators and automated measures, comparing RAG-based responses to human ground truth and the existing BERT-based system (Figure \ref{fig:system_overview} shows an overview of traditional customer care scenario with existing and proposed system). 
In short, we answer the following research questions.

\begin{enumerate}
    \item \textbf{RQ1:} What are the effects of different embedding techniques, retrieval strategies, and prompting methods on RAG performance?
    \item \textbf{RQ2:} Do RAG-based responses provide greater assistance to human agents compared to the existing BERT-based  system?
    \item \textbf{RQ3:} Can the ReAct (Reason+Act) prompting improve factual accuracy and reduce hallucinations in LLM in real-time settings?
    
\end{enumerate}

Our findings demonstrate an overall improvement over the existing system by suggesting more accurate and relevant responses, highlighting the potential of RAG LLM as an excellent choice for customer care automation.

\begin{figure*}[h]
    \includegraphics[width=\textwidth, height=6cm]{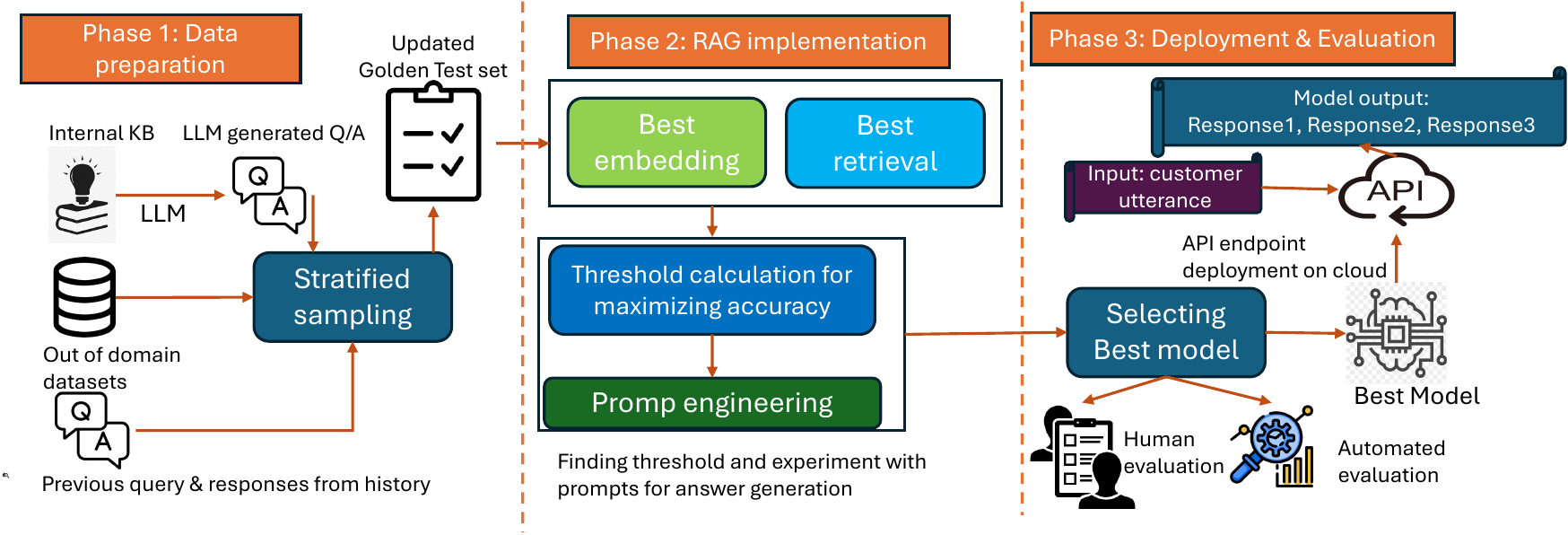}
    \caption{End to end RAG LLM framework }
     \label{fig:rag-end-to-end}
\end{figure*}

\section{Related Work}

\paragraph{RAG architecture:}
RAG has emerged as a promising solution by incorporating knowledge from external databases to overcome the hallucination, outdated knowledge, transparency issues for LLMs \cite{lewis2021retrievalaugmented,khandelwal2020generalization,borgeaud2022improving,JMLR:v24:23-0037,yasunaga2023retrievalaugmented}. Traditional RAG, popularized after the adoption of ChatGPT, follows a simple process of indexing, retrieval, and generation \cite{gao2024retrievalaugmented} .
Despite advancements in Advanced and Modular RAG, Traditional RAG remains popular in the industry due to its ease of development, integration, and quicker speed to market \cite{liu2023lost}. The core components of Traditional RAG include Retriever, Generator, and Augmentation Method, with research focusing on improving semantic representation \citep{dai2022promptagator,zhang2023retrieve}, query alignment \cite{wang2023knowledgpt}, and integration with LLMs \cite{yu2023augmentationadapted,shi2023replug,izacard2022atlas}, which motivate our RQ1 to find the optimum setup in this specific use case of RPS in the contact center.

\paragraph{RAG LLM for question answering:}
Several open-domain question-answering (Q/A) tasks have been completed by RAG-based architectures efficiently \cite{siriwardhana2023improving, zhao2024retrieval, izacard2020leveraging}. With the advent of LLMs in recent periods, multiple studies also focus on utilizing LLMs for customer assistance, specifically in recommendations \cite{yang2023palr,friedman2023leveraging} and dialogue generation \cite{shuster2022blenderbot, zhang2023memory}. Recent work by \cite{baek2024knowledge} proposes augmenting LLMs with user-specific context from search engine interaction histories to personalize outputs, leveraging entity-centric knowledge stores derived from users' web activities. Similarly, \cite{xu2024retrieval} introduces a customer service question-answering approach integrating RAG with a knowledge graph (KG) constructed from historical issue data. Therefore, our study is motivated by these prior researches to integrate RAG as a retrieval tool and utilize LLM to generate responses to answer customer queries.

\section{Methodology}



To implement an end-to-end RAG framework with LLM, first, it is essential to create a comprehensive dataset comprising relevant question-answer pairs along with corresponding knowledge documents. Next, design choices for specific components of the RAG and LLM architecture must be finalized. Finally, the model should be thoroughly evaluated and refined before being deployed into production.

\subsection{Phase I: Data Preparation} \label{DataPrep}
An ideal golden dataset for evaluating RAG architecture (Figure \ref{fig:rag-end-to-end}) should include:
\begin{enumerate}
    \item Domain-specific questions (previous queries) with their corresponding grounded responses.
    \item Relevant knowledge base (KB) articles (company documents) containing the policies that determine answers to specific queries.
    \item Out-of-domain questions to ensure the LLM can handle generic queries without hallucinating and can guide customers to provide relevant queries.
\end{enumerate}

\begin{table}[h]
\centering
\resizebox{0.5\textwidth}{!}{
\begin{tabular}{@{}l|l|l|l@{}}
\toprule
\textbf{Source}                                                                                       & \textbf{Total \#} & \textbf{length (query)} & \textbf{length (response)} \\ \midrule
\textbf{KB articles}                                                                                  &            1205       & \multicolumn{2}{l}{Avg. document length: 134.25}            \\ \hline
\textbf{\begin{tabular}[c]{@{}l@{}}In domain Q/A (generated\\ by LLM from KB articles)\end{tabular}}  &      4785             &            10.7             &          33.59                  \\ \hline
\textbf{\begin{tabular}[c]{@{}l@{}}In domain Q/A (sampled\\ from previous history)\end{tabular}}      &        3000           &               9.58          &             28.81               \\ \hline
\textbf{\begin{tabular}[c]{@{}l@{}}Out domain Q/A (sampled\\ from MS-MARCO)\end{tabular}}             &     3660              &            5.73             &          5                  \\  \bottomrule
\end{tabular}
}
\caption{Total number and average length (in terms of word count) for the KB articles and various Q/A pairs for the RAG implementation}
\label{tab_dataset_statistics}
\end{table}

To create a robust test set (Table \ref{tab_dataset_statistics} for details), we utilize LLM to generate both relevant question-answer pairs from the company's KB articles (refer to Section A in the Appendix for prompts). Additionally, we supplement these relevant pairs with samples from previous queries \& responses in the contact center with out-of-domain questions by sampling from open-source datasets such as MS-MARCO \cite{bajaj2018ms}.

\subsection{Phase II: RAG}

The main components of the RAG architecture are the Retriever and the Generator LLM. We evaluate various strategies for each component and finalize our choices for production. Our findings are validated through experiments with several open-domain question-answer datasets, including MARCO \cite{bajaj2018ms}, SQuAD \cite{rajpurkar-etal-2016-squad}, and TriviaQA \citep{joshi-etal-2017-triviaqa} (details in Appendix).  
They present a comparable level of question-answering challenges where answers can be derived from the retrieved knowledge base.

\textbf{Embedding Strategy:}
The best embedding strategy ensures high performance of the retriever and affects downstream tasks like response generation. We compare the Universal Sentence Encoder (USE) embeddings \cite{cer2018universal}, Google's Vertex AI embedding model text-embedding-gecko@001 \cite{anil-etal-2022-large}, and SBERT-all-mpnet-base-v2 \cite{reimers2019sentencebert} from the sentence-transformers collection.


\textbf{Retrieval strategies: } By retrieving relevant passages from a large corpus of KB articles, the model gains crucial contextual information, enhancing response accuracy and coherence. 
We specifically consider ScaNN \cite{google-research2020scann} for its efficiency in handling large-scale datasets and KNN HNSW \cite{malkov2018efficient} for its efficient memory usage as retrieval strategies in our study. 
Additionally, we tested different retrieval thresholds to ensure incorrect documents are not retrieved and passed to the LLM for response generation.

\textbf{LLM for generation: }
Once the best embedding strategy, retrieval technique, and retrieval thresholds are identified, we test different prompting techniques to ensure that LLMs generate grounded factual responses. We utilize PaLM2 foundation models (text-bison, text-unicorn) \cite{anil2023palm} for text generation across all tasks, as they offer a clear path to production in terms of enterprise licenses and security requirements with Google’s models, compared to other available LLMs at the time of our research.

The best model from Phase 2, incorporating the optimal embedding, retrieval, and prompting techniques, is packaged with relevant KB articles and deployed on a cloud Virtual Machine. For real-time usage, an endpoint is created that takes a customer query and the conversation context as input, generating response suggestions as output.

\section{Results and Findings}


First, we optimize RAG setup for our use cases, then evaluate LLM responses using automated metrics and human evaluations. Finally, we assess if prompting or ReAct strategies can improve real-world performance to an acceptable level.

\subsection{Retrieval evaluation}
\paragraph{Best setting for RAG}:

We assessed retriever efficiency using the "Recall at K" ($R@k$) metric, where $K$ represents the top 1, 3, 5, or 10 documents retrieved, measuring how well the retriever retrieves relevant documents. 
The Vertex AI - textembedding-gecko@001 (768) embedding, paired with ScaNN retrieval, yielded the best outcomes. Overall, ScaNN generally outperformed KNN HNSW in most cases due to its efficient handling of large-scale datasets and superior retrieval accuracy through quantization and re-ranking techniques \cite{chang2024pefa}, so we include only the ScaNN results in Table \ref{tab_rag_best_combo}. Similarly, Vertex AI embeddings  surpassed Sentence BERT and USE due to its superior ability to capture complex semantic relationships tailored for large-scale industry applications.

\begin{table}[h]
\centering
\footnotesize
\resizebox{0.5\textwidth}{!}{
\begin{tabular}{@{}lcccc@{}}
\toprule
\textbf{Embedding} & \multicolumn{1}{l}{\textbf{Embedding size}} & \multicolumn{1}{l}{\textbf{R@1}} & \multicolumn{1}{l}{\textbf{R@3}} & \multicolumn{1}{l}{\textbf{R@5}} \\ \midrule
\textbf{USE}       & 512                                         & -                                & -                                & -                                \\
\textbf{SBERT}     & 768                                         & +15.36                           & +9.42                            & +8.22                            \\
\textbf{Vertex AI} & 768                                         & +21.55                           & +13.87                           & +11.85                           \\ \bottomrule
\end{tabular}
}

\caption{Performance of different embedding technique in the company data for ScaNN. Performance is shown as the (\%) of improvement wrt the lowest performing embedding (USE in this scenario)}
\label{tab_rag_best_combo}
\end{table}

\begin{figure}
    \includegraphics[width=6cm, scale = 0.8]{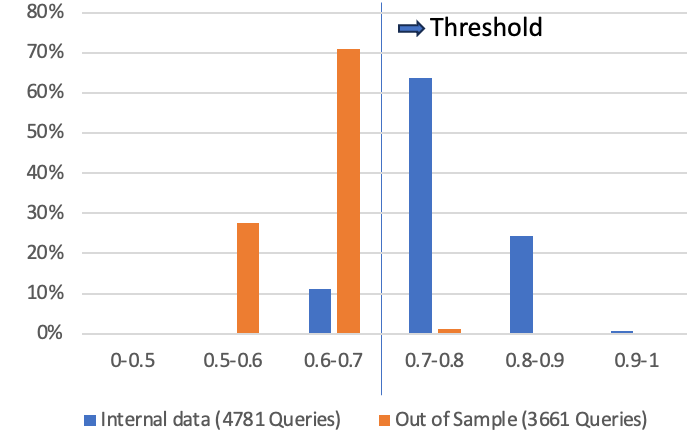}
    \caption{Cosine similarity score between query and ScaNN retrieved Document; retrieval threshold(0.7)}
     \label{fig:retrieval threshold}
\end{figure}

\paragraph{Retrieval Threshold:}
For out-of-domain or trivial customer queries like "Hello" or "Bye," document retrieval is unnecessary, as shown by 98.59\% of retrieved articles having a cosine similarity score below 0.7. In contrast, 88.96\% of articles retrieved for relevant company data questions scored above 0.7 (Figure \ref{fig:retrieval threshold}). This suggests that setting the retrieval threshold at 0.7 effectively determines when retrieval is needed, thereby enhancing response generation efficiency.

\subsection{Response Prediction System Evaluation} \label{Evaluation Section}
To develop an effective Response Generation System (RPS), we conducted a comprehensive evaluation comparing RAG LLM-based responses with a current BERT-based algorithm. Using 1,000 real contact center chat transcripts (PII and PCI compliant), comprising over 5,000 messages, we analyzed customer queries, human agent responses, RAG LLM suggestions, BERT-based suggestions, and retrieved knowledge base documents to assess quality, consistency, and factuality through automated measures and human evaluations.

\begin{table}[]
\centering
\footnotesize
\begin{tabular}{@{}ll@{}}
\toprule
\textbf{Metric}              & \textbf{Improvement} \\ \midrule
\textbf{Accuracy}            &          +10.15            \\
\textbf{Hallucination}       &             -4.76         \\
\textbf{Missing rate}        &            -5.43          \\
\textbf{AlignScore}          &             +5.6         \\
\textbf{Semantic similarity} &             +20.01         \\
\textbf{AI-generated}        &             -40.17         \\ \bottomrule
\end{tabular}
\caption{Comparision between RAG based responses and existing BERT-based ones for automated evaluations. Values indicate the difference in percentage (\%) as average of all samples}
\label{tab:automated_score}
\end{table}

\subsubsection{Automated evaluations}
\label{subsec-automated_evaluations}
We utilize the following evaluation techniques, with Table \ref{tab:automated_score} illustrating our RAG LLM-based technique's performance against the current BERT-based system.

\paragraph{Accuracy, Hallucination and Missing rate evaluation}
In a question-answer system, a response to each query can generate one of three types of responses: accurate (correctly answers the question), hallucinate (incorrect answer), or missing (no answer generated). Therefore, our approach, inspired by \cite{sun2023headtotail} which provides 98\% agreement with human judgments, utilizes an LLM-based method. We employ ChatGPT-3.5-turbo as our evaluator LLM.
We prompted the LLM with a query, generated response, and original human response, categorizing the LLM's responses as "correct" for factual and semantic alignment, "incorrect" for mismatches, and "unsure" for semantic challenges. Evaluation includes Accuracy (correct responses), Hallucination Rate (incorrect responses), and Missing Rate (unsure responses) metrics as the proportion of corresponding responses. Overall, RAG LLM improves accuracy by reducing hallucinations and missing rates compared to BERT responses.

\paragraph{AlignScore:}
To ensure response alignment with KB articles, we use AlignScore \cite{zha2023alignscore} to measure information consistency. Evaluating RAG LLM and BERT-based models on utterances with relevant KB article retrieval by RAG, RAG LLM shows a statistically significant 5.6\% improvement via Student's t-test. This enhancement derives from integrating retrieved documents as prompts for LLM responses, whereas BERT relies on query-answer pairs in its training dataset.

\paragraph{Semantic similarity:}
To ensure usability by human agents, coherence between generated and original human responses is crucial. We measure semantic similarity using LongFormer embeddings \cite{beltagy2020longformer}, calculating cosine similarity between generated and original human responses for both models. RAG LLM exhibits an average 20\% higher similarity, a statistically significant improvement.

\paragraph{Human touch:}
Customer service are typically preferred to be handled by humans \citep{fernandes2021understanding,wu2022tech}, emphasizing the importance of generating human-like responses. We use the AI text detector GPTZero \citep{GPTzero}, with a 99.05\% true positive rate for human responses in our dataset, to evaluate response naturalness. Assessing AI percentage (utterances identified as AI-generated), the BERT-based system, which selects responses from human-generated options, sounds more human.

\subsubsection{Human evaluations}
Our method aims to support rather than replace humans through a human-in-the-loop approach. We thoroughly evaluate the quality of RAG LLM and BERT responses using human annotators. Each response is assessed against several criteria, and the average score is computed from all annotators' evaluations. Evaluation metrics were grouped into three main categories:
\paragraph{Human Preference Score:} Following the classical approach of which version humans prefer most  \cite{wu2023human, stiennon2020learning}, we evaluated which model's responses—"BERT" or "RAG"—were preferred by human evaluators.

\paragraph{Quantitative Metrics:} Similar to \cite{sun2023headtotail}, we evaluated factual accuracy (based on human judgment of 'correct,' 'incorrect,' or 'unsure'). Accuracy, Hallucination, and Missing rates were calculated as the number of correct, incorrect, and unsure responses divided by the total number of responses evaluated, respectively.

\paragraph{Qualitative Metrics:} 
\begin{enumerate}
    \item \textbf{Contextual Relevance:} Assessed whether the predicted responses were appropriate and in line with the  context of the conversation.
   \item \textbf{Completeness:} Checked if the predicted responses were fully-formed and could be used as complete answers by the agents in specific parts of the conversation.
\item \textbf{Specificity:} Determined whether the predicted responses were tailored to the specific conversation or were too general. Human annotators scored these metrics on a scale of 0 (lowest) to 2 (highest).
\end{enumerate}

The results, as detailed in the Table \ref{table: Generation metrics for SR overall}, Responses generated by the RAG model demonstrated a 45\% improvement in factual accuracy and a 27\% decrease in the rate of hallucinations compared to the existing model. Moreover, the human evaluators favored responses from the RAG model over the current production model 75\% of the times. 

The Response Prediction System was deployed using Flask, a standard micro web framework, and Gunicorn, chosen for its performance, flexibility, and simplicity in production system configuration. The API receives customer queries as input and provides answers as output. The API was thoroughly load tested using Locust, an open-source performance/load testing tool for HTTP and other protocols, to ensure it meets real-time latency requirements in a production setting. Finally, the API was integrated with the Agent Workspace UI to deliver predictions to Contact Center agents, assisting customers in real-time.

\begin{table}\small
\centering
\begin{tabular}{lllll}
\toprule
\textbf{Metric} & \textbf{Improvement} \\ \midrule
\textbf{Contextual   Relevance} & +48.14 &  &  &  \\
\textbf{Specificity}            & +97.97 &  &  &  \\
\textbf{Completeness}           & +70.15 &  &  &  \\
\textbf{Accuracy}               & +45.69 &  &  &  \\
\textbf{Hallucination   Rate}   & -27.49 &  &  &  \\
\textbf{Missing Rate}           & -70.02 &  &  &  \\
\textbf{Preference}           & +200.61 &  &  &  \\
\bottomrule
\end{tabular}
\caption{
Human evaluation comparison (\% diff.) between RAG and existing BERT-based ones.
}
\label{table: Generation metrics for SR overall}
\end{table}

\subsection{Evaluation for ReAct and prompting techniques}

\paragraph{Experiments with ReAct}
To answer our third RQ, we utilized ReAct Tools to determine when to activate the information retrieval component within the RAG framework, while maintaining the same retrieval, embeddings, and generation strategies. We evaluated two scenarios: "RAG with ReAct" and "RAG without ReAct," with K = 3. As shown in Table \ref{table: ReAct_accuracy_hallucination_vs_baseline}. While ReAct improved the accuracy by 7\% and reduced hallucination by 13.5\%, it resulted in slower performance \ref{table: ReAct_latency}, making it inconvenient in real-time conversation.

\begin{table}\scriptsize
\centering
\begin{tabular}{clccc}
\hline
\textbf{K} & \textbf{Strategy} & \textbf{Accuracy} & \textbf{Hallucination Rate} & \textbf{Missing Rate} \\
\hline
1          & \textbf{ReAct}                 & -2.13           & +51.40                     & -34.38               \\
3          & \textbf{ReAct}                 & +7.08           & -13.48                     & -19.38              \\
3 & \textbf{CoVe} & -43.65           & +27.43                       & +11.35                \\
3 & \textbf{CoTP}  & -3.45           & +1.33                       & +1.98      \\
\hline
\end{tabular}
\caption{
Comparison (\% diff.) of ReAct RAG (with different values of k), CoVe and CoTP performance with respect to baseline on company data.
}
\label{table: ReAct_accuracy_hallucination_vs_baseline}
\end{table}

\begin{table}[h!]\scriptsize
\centering
\begin{tabular}{ccc}
\hline
\textbf{RAG Strategy  }            & \textbf{95th Percentile} & \textbf{99th Percentile} \\
\hline
reAct & 4.0942           & 6.2084           \\
non-reAct      & 0.8850           & 1.1678   \\
\hline
\end{tabular}
\caption{
Latency Comparison (seconds) between ReAct RAG and non-ReAct RAG based on 10000 queries
}
\label{table: ReAct_latency}
\end{table}

\paragraph{Prompting Techniques Experiments}
We evaluated Chain of Verification (CoVe) \citep{dhuliawala2023chainofverification} and Chain of Thought Prompting (CoTP) \cite{wei2023chainofthought} to improve factual accuracy and reduce hallucinations. However, both techniques are time-consuming, requiring multiple LLM calls per query, and did not show significant improvements for the Company data. CoVe was 43\% less accurate and CoTP was 3\% less accurate (see Table \ref{table: ReAct_accuracy_hallucination_vs_baseline}). Therefore, we decided against using these prompting techniques.

\section{Conclusion}
In this study, we demonstrate the practical challenges of implementing a RAG-based Response Prediction System in an industry setting. We evaluated various retrieval and embedding strategies combined with different prompting techniques to identify the best combinations for different use cases. Our evaluations show that retrieving relevant knowledge base articles and generating responses from LLMs can be more contextually relevant and accurate than BERT responses, which choose from the most relevant query-answer pairs. We also highlight that ReAct and advanced prompting techniques may not be practical for industry settings due to latency issues. Overall, our approach indicates that implementing RAG-based LLM response generation for contact centers is feasible and can effectively aid humans, reducing their workload. In the future, we plan to advance our work in three directions. Firstly, we aim to evaluate other LLMs. Secondly, we will test if query rewriting and reformulation can improve retrieval performance. Lastly, we intend to explore advanced RAG approaches to integrate knowledge bases from various sources.

\section*{Limitations}
Despite ongoing significant research, LLMs remain unpredictable. Though this paper showcases work on grounding the generated responses, LLMs to a certain degree are still capable of generating inaccurate information based on their learnt parametric memory. This work also does not focus on other LLM issues such as context length constraints, prompt injections and quality of Knowledge base data. Addressing other open challenge such as biases along with their ethical consideration are also not considered in the scope of this paper. The paper also does not address or evaluate RAG for multilingual data sources. 

\section*{Ethics Statement}
The data set used for training and validation of LLMs in this paper do not have any unbalanced views or opinions of individuals that might bias the generated response. The LLMs are still capable of generating inaccurate responses based their parametric memory even when relevant contextual information might be provided. Filtering toxic responses and prompt injections are also not considered in this evaluation. 

\bibliographystyle{ACM-Reference-Format}
\bibliography{main}

\appendix

\section{Evaluation of RAG Approach for Open Source Datasets} 
\label{sec:appendix}

To evaluate the effectiveness of the RAG-based approach, we also conducted a sample study using three open-domain datasets following a similar methodology.

\subsection{Dataset Statistics}
We consider three popular open source question answering datasets focusing on a varitey of topics. A random subset of  \textbf{MS MARCO}\citep{bajaj2018ms}, \textbf{SQuAD} \citep{rajpurkar-etal-2016-squad}, \textbf{TriviaQA} \citep{joshi-etal-2017-triviaqa} were considered for the evaluation. Table \ref{table: Dataset statistics} shows the brief overview of the considered datasets.

\begin{table}[h]\tiny
\centering
\resizebox{0.48\textwidth}{!}
{
\begin{tabular}{c|c|c|c|c|c|c}
\hline
Dataset  & \vtop{\hbox{\strut Unique}\hbox{\strut Docs}} & \vtop{\hbox{\strut Ave.}\hbox{\strut Doc.}\hbox{\strut Tokens}}   & \vtop{\hbox{\strut Unique}\hbox{\strut Quest-}\hbox{\strut ions}} & \vtop{\hbox{\strut Ave.}\hbox{\strut Question}\hbox{\strut Tokens}} & \vtop{\hbox{\strut Unique}\hbox{\strut Answers}} & \vtop{\hbox{\strut Ave.}\hbox{\strut Answer}\hbox{\strut Tokens}} \\
\hline
MS-MARCO & 4997             & 58.80                 & 5000             & 5.67                   & 4999           & 14.37               \\
SQUAD    & 5000             & 94.21                 & 4994             & 10.33                 & 4413           & 2.59                \\
TRIVIA   & 3530             & 4321.4                  & 4087             & 12.95                   & 3471           & 1.67   \\
\hline
\end{tabular}}
\caption{
Open source dataset random sample statistics
}
\label{table: Dataset statistics}
\end{table}

\subsection{Retrieval and Embedding Performance}
We observed a specific trend in Recall values in lower k values (1, 3) versus higher k values (5, 10) for SQuAD and TRIVIA. For SQuAD, Vertex AI textembedding-gecko@001 (768) embedding with ScaNN retrieval performed the best at lower k but at higher k, SBERT-all-mpnet-base-v2 (768) with ScaNN performed better. For TRIVIA, SBERTall-mpnet-base-v2 (768) embedding with HNSW KNN retrieval performed the best at lower k but at higher k, SBERT-all-mpnet-base-v2 (768) with ScaNN performed better. For MSMARCO, Vertex AI -textembedding-gecko@001 (768) embedding with ScaNN retrieval combination was a clear winner. Refer Table \ref{table: Retrieval Metrics for open source data sets} for more details.

\begin{table*}[t]
\centering
\resizebox{\textwidth}{!}{
\begin{tabular}{c|c|c|c|c|c|c}
\hline
\textbf{Dataset} &
  \textbf{Embedding Strategy} &
  \textbf{Retrieval Strategy} &
  \textbf{Recall @ 1} &
  \textbf{Recall @ 3} &
  \textbf{Recall @ 5} &
  \textbf{Recall @ 10}\\
\hline
SQUAD                & USE (512)                                 & HNSW KNN & 0.4188    & 0.5946               & 0.6634               & 0.7424               \\
SQUAD                & SBERT - all-mpnet-base-v2 (768)           & HNSW KNN & 0.6708    & 0.8444               & 0.8902               & 0.9336               \\
SQUAD                & Vertex AI - textembedding-gecko@001 (768) & HNSW KNN & 0.6958    & 0.8486               & 0.8804               & 0.911                \\
SQUAD                & USE (512)                                 & ScaNN                & 0.4282    & 0.6116               & 0.6834               & 0.7666               \\
SQUAD                & SBERT - all-mpnet-base-v2 (768)           & ScaNN                & 0.685     & 0.8636               & 0.913                & 0.9584               \\
SQUAD                & Vertex AI - textembedding-gecko@001 (768) & ScaNN                & 0.7156    & 0.874                & 0.908                & 0.9414               \\
TRIVIA               & USE (512)                                 & HNSW KNN & 0.459     & 0.6004               & 0.6604               & 0.7333               \\
TRIVIA               & SBERT - all-mpnet-base-v2 (768)           & HNSW KNN & 0.793     & 0.8691               & 0.8921               & 0.9171               \\
TRIVIA               & Vertex AI - textembedding-gecko@001 (768) & HNSW KNN & 0.6423    & 0.7487               & 0.782                & 0.8233               \\

TRIVIA               & USE (512)                                 & ScaNN                & 0.4101    & 0.6039               & 0.6687               & 0.7548               \\
TRIVIA               & SBERT - all-mpnet-base-v2 (768)           & ScaNN                & 0.7086    & 0.8654               & 0.8992               & 0.9288               \\
TRIVIA               & Vertex AI - textembedding-gecko@001 (768) & ScaNN                & 0.5936    & 0.759                & 0.8038               & 0.8537               \\
MS-MARCO             & USE (512)                                 & HNSW KNN & 0.5263    & 0.6856               & 0.7347               & 0.784                \\
MS-MARCO             & SBERT - all-mpnet-base-v2 (768)           & HNSW KNN & 0.9128    & 0.9798               & 0.9878               & 0.9925               \\
MS-MARCO &
  Vertex AI - textembedding-gecko@001 (768) &
  HNSW KNN &
  0.8194 &
  0.9241 &
  0.9425 &
  0.9577 \\
MS-MARCO             & USE (512)                                 & ScaNN                & 0.5347    & 0.6996               & 0.7518               & 0.8035               \\
MS-MARCO             & SBERT - all-mpnet-base-v2 (768)           & ScaNN                & 0.9132    & 0.9816               & 0.9896               & 0.9944               \\
MS-MARCO             & Vertex AI - textembedding-gecko@001 (768) & ScaNN                & 0.8296    & 0.9376               & 0.9581               & 0.9738               \\
\hline
    
\end{tabular}}
\caption{
Recall@K for retrieval and embedding strategies for different data sets. 
}
\label{table: Retrieval Metrics for open source data sets}
\end{table*}

\subsection{Evaluation of Generated Responses}
Table \ref{table: Generation metrics for data sets} shows the accuracy, hallucination, and missing rate for the open sources datasets (through automated evaluation as described in Subsection \ref{subsec-automated_evaluations}. As observed in the Retrieval evaluation section, we notice a similar relationship in the accuracy and hallucination as document size increases. From Table \ref{table: Dataset statistics}, we observe that the token length of the TriviaQA documents is much larger than MS-MARCO and SQuAD. Similarlt, we observed lower accuracy and higher hallucination rates with TriviaQA when compared to MS-MARCO and SQuAD. 
Table \ref{table: CoTP open source datasets} and \ref{table: CoVe open source datasets} show performance of Chain of Thought Prompting and Chain of Verification performance on  open-source datasets.
Accuracy and hallucination rate improvement vary based on the open source dataset.

\begin{table}\scriptsize
\centering
\begin{tabular}{ccccc}
\hline
\textbf{k} &
  \textbf{Data} &
  \textbf{Accuracy} &
  \textbf{Hallucination} &
  \textbf{Missing} \\
    \ &
  \ &
  \ &
  \textbf{rate} &
  \textbf{rate} \\
  \hline
1 &   SQUAD    & 84.74 & 8.36  & 6.9  \\
1 &   TriviaQA & 58.48 & 28.68 & 12.8 \\
1 &   MS-MARCO        & 89.06 & 7.06  & 3.86 \\
3 &   SQUAD    & 91.32 & 5.48  & 3.2  \\
3 &   TriviaQA & 25.08 & 67.41 & 7.46 \\
3 &   MS-MARCO        & 89.9  & 6.93  & 3.16 \\

\hline

\end{tabular}

\caption{
Generation quality metrics (\%) using text-bison@001 and ScaNN 
}
\label{table: Generation metrics for data sets}
\end{table}

\begin{table}[]\tiny
\begin{tabular}{ccccccc}
\hline
\multirow{2}{*}{\textbf{dataset}} & \multicolumn{2}{c}{\textbf{accuracy}} & \multicolumn{2}{c}{\textbf{hallucination\_rate}} & \multicolumn{2}{c}{\textbf{missing\_rate}} \\

                      & \textbf{baseline}       & \textbf{CoTP}        & \textbf{baseline}            & \textbf{CoTP}             & \textbf{baseline}         & \textbf{CoTP}           \\
                      \hline
SQUAD                 & 98.14         & 94.58      & 1                & 4.16           & 0.86           & 1.9          \\
TRIVIA                & 63.95         & 86.27      & 22.85              & 6.45           & 13.18           & 6.38         \\
MS-MARCO              & 92.1          & 90.94      & 4.64              & 4.32           & 3.26           & 4.74        \\
\hline
\end{tabular}
\caption{
Baseline vs CoTP evaluation metrics (\%) without retrieval (question-doc. pair used as it is)
}
\label{table: CoTP open source datasets}
\end{table}

\begin{table}[]\tiny
\begin{tabular}{ccccccc}
\hline
\multirow{2}{*}{\textbf{dataset}} & \multicolumn{2}{c}{\textbf{accuracy}} & \multicolumn{2}{c}{\textbf{hallucination\_rate}} & \multicolumn{2}{c}{\textbf{missing\_rate}} \\

                      & \textbf{baseline}       & \textbf{CoVe}        & \textbf{baseline}            & \textbf{CoVe}             & \textbf{baseline}         & \textbf{CoVe}           \\
\hline
SQUAD                 & 98.14         & 95.96      & 1               & 3.24           & 0.86           & 0.8          \\
TRIVIA                & 63.95         & 63.76      & 22.85              & 23.83           & 13.18           & 12.4         \\
MS-MARCO              & 92.1          & 92.6       & 4.64              & 5.54           & 3.26           & 1.86        \\
\hline
\end{tabular}
\caption{
Baseline vs CoVe evaluation metrics (\%) without retrieval (question-doc. pair used as it is)
}
\label{table: CoVe open source datasets}
\end{table}

\section{Prompt Examples}
We utilize LLMs for various tasks in our methodology, including question-answer pair generation from knowledge base articles, response generation, factual accuracy evaluation, and advanced CoTP \& CoV prompts. Therefore, we include the specific prompts used for these different tasks.

\paragraph{Prompt for answer generation} 

You are a reading comprehension and answer generation expert. Please answer the question from the document provided. If the document is not related to the question, simply reply: "Sorry, I cannot answer this question". Following are the guidelines you need to follow for generating the responses:

1) They should always be professional, positive, friendly, and empathetic.
2) They should not contain words that have a negative connotation (Example: "unfortunately").
3) They should always be truthful and honest.
4) They should always be STRICTLY less than 30 words. If the generated response if greater than 30 words, rephrase and make it less than 30 words.

document: <retrieved\_document>,

question: <question>, 

output: 

\paragraph{Prompt for Hallucination Judgement}:

    You need to check whether the prediction of a question-answering systems to a question is correct. You should make the judgement based on a list of ground truth answers provided to you. You response should be "correct" if the prediction is correct or "incorrect" if the prediction is wrong. Your response should be "unsure" where there is a valid ground truth and prediction is "Sorry, I don't know." or if you are not confident if the prediction is correct. 
    
    Below are the different cases possible:
    
    1) Examples where you should return "correct".
    
    Question: What is the customer registration process? 
    
    Ground Truth: The customer registration process is a way for customers to create an account with them. This allows them to track their purchases, receive personalized offers, and more. The process is simple and can be completed in a few minutes. 
    
    Prediction: The customer registration process is a process that allows customers to register their information with them. This process allows customers to receive benefits such as discounts, special offers, and personalized shopping experiences.
    
    Correctness: correct
    
    Question: What happens if my refund is pending?
    
    Ground Truth: Sorry, I don't know.
    
    Prediction: Sorry, I don't know.
    
    Correctness: correct
    \newline
    2) Examples where you should return "incorrect".
    
    Question: What do I need to do to get the military discount?
    
    Ground Truth: You need to have a smartphone and be registered for the discount. If you don't have a smartphone, you can use discount code RC5. If you are in the pilot 425 stores area, you can key in your phone number.
    
    Prediction: The military discount is available to active duty military members, veterans, and their families. The discount is 10 percent off eligible purchases.
    
    Correctness: incorrect
    
    Question: How do I apply for the consumer card?
    
    Ground Truth: Sorry, I don't know.
    
    Prediction: You can apply for the consumer card in-store, online or by mail.
    
    Correctness: incorrect
    
    3) Examples where you should return "unsure".
    
    Question: What is the Return Policy?
    
    Ground Truth: The Return Policy is available on the website. You can find it by searching for "Return Policy" or by clicking on the link in the article.
    
    Prediction: Sorry, I don't know.
    
    Correctness: unsure

    Provide correctness for the below question, ground truth and prediction: 
    
    Question: <question>
    
    Ground Truth: <ground truth>
    
    Prediction: <prediction>
    
    Correctness:

\textbf{Prompt for Chain of Prompting}:
Prompt for quote extraction: You are a reading comprehension and quote extraction expert. Please extract, word-for-word, any quotes relevant to the question. If there are no quotes in this document that seem relevant to the provided question, please say "I can’t find any relevant quotes".

For document: <document>,

question: <question>, 

output:

\paragraph{Prompt for Generating Baseline Response and Plan Verification (Chain of Verification)}:

    Below is a question:
    <question>

    Below is the document from which the answer should be generated:
    <document>

    You are an subject matter expert working at Contact Centers. Your expertise includes quote extraction, answer generation, and asking verification questions to improve the overall factual accuracy of the answers you provide.

    Your first goal is to extract, word-for-word, any quotes relevant to the question that could be used to answer the question. If there are no quotes in this document that seem relevant to the provided question, simply return: "I can't find any relevant quotes".
    
    Your second goal is to use *solely* the quotes extracted from the first goal and generate a concise and accurate answer (using the below listed guideline) by rephrasing the quotes to answer the question. If the quotes could not be used to answer the question, simply return: "Sorry, I cannot answer this question".
        1) They should always be professional, positive, friendly, and empathetic.
        2) They should not contain words that have a negative connotation (Example: "unfortunately").
        3) They should always be truthful and honest.
        4) They should always be STRICTLY less than 30 words. If the generated response if greater than 30 words, rephrase and make it less than 30 words.

    Your third goal is to generate a list of potential areas that might require verification based on the content of the document to increase factual accuracy of the answer. Your response should be in the below format:

    ```
    Quotes: <Your Extracted Quotes>
    
    Answer: <Your Answer>

    Potential Areas for Verification:
    1) Your Specific point or segment from your answer.
    2) Your Another point or segment from your answer.
    N) Your Nth point or segment from your answer. 
    ```

\paragraph{Prompt for Executing Verification Questions and Generating Verified Response (Chain of Verification)}:

    Below is a question:
    <question>

    Below is the answer:
    <answer>

    Below is the document from which the answer was generated:
    <document>

    Based on the potential areas for verification:
    <areas of verification>
    
    You are an subject matter expert working at Contact Centers. Your expertise includes improvising answers to questions about the company to increase factual correctness using the factual accuracy verification questions provided to you.

    Your goal is to check each verification point against the document, provide feedback on any inconsistencies, and then generate a final verified (using the below listed guidelines), concise and accurate answer in strictly less than 30 words that addresses the factual inconsistencies. 
        1) They should always be professional, positive, friendly, and empathetic.
        2) They should not contain words that have a negative connotation (Example: "unfortunately").
        3) They should always be truthful and honest.
        4) They should always be STRICTLY less than 30 words. If the generated response if greater than 30 words, rephrase and make it less than 30 words.
    
    Your response should be in the below format:

    ```
    Feedback:
    1) Your Verification for point 1.
    2) Your Verification for point 2.
    N) Your Verification for point N.

    Final Verified Response: [Your Revised Response]
    ```

\paragraph{Prompt for generating answer from a document and question (open source datasets).}

    You are a question answering bot.
    Your job is to generate answer to the question using the provided articles. 
    The answers should be derived only from the articles. If the answer is not present in the articles, return the text - NOANSWERFOUND.
    The answer should be less than 10 words and in a sentence format.

    Example where answer could not be found in the articles:
    
    Question: Which county is Smyrna city in?
    
    Document: Georgia is a southeastern U.S. state whose terrain spans coastal beaches, farmland and mountains. Capital city Atlanta is home of the Georgia Aquarium and the Martin Luther King Jr. National Historic Site, dedicated to the African-American leader’s life and times. 
    
    Return Text: NOANSWERFOUND

    Example where answer could be found in the articles:
    
    Question: Which county is Smyrna city in?
    
    Document: Smyrna is a city in Cobb County, Georgia, United States. Cobb County is a county in the U.S. state of Georgia, located in the Atlanta metropolitan area in the north central portion of the state.
    
    Return Text: Cobb County of the state of Georgia

    Provide answer to the below Question/Query using the below Document.

    Question: <question>
    
    Document: <document>
    
    Return Text:









\end{document}